\documentclass[10pt,twocolumn,letterpaper]{article}

\usepackage{cvpr}
\usepackage{times}
\usepackage{epsfig}
\usepackage{graphicx}
\usepackage{amsmath}
\usepackage{amssymb}
\usepackage{subfigure}
\usepackage{multirow}

\usepackage[breaklinks=true,bookmarks=false]{hyperref}

\cvprfinalcopy 

\def\cvprPaperID{****} 

\begin{document}

\title{YH Technologies at ActivityNet Challenge 2018}

\author{Ting Yao and Xue Li\\
         YH Technologies Co., Ltd, Beijing, China\\
                                {\tt\small \{tingyao.ustc, miya.lixue\}@gmail.com}
}

\maketitle

\begin{abstract}
This notebook paper presents an overview and comparative analysis of our systems designed for the following five tasks in ActivityNet Challenge 2018: temporal action proposals, temporal action localization, dense-captioning events in videos, trimmed action recognition, and spatio-temporal action localization.

\textbf{Temporal Action Proposals (TAP)}: To generate temporal action proposals from videos, a three-stage workflow is particularly devised for TAP task: a coarse proposal network (CPN) to generate long action proposals, a temporal convolutional anchor network (CAN) to localize finer proposals, and a proposal reranking network (PRN) to further identify proposals from previous stages. Specifically, CPN explores three complementary actionness curves (namely point-wise, pair-wise, and recurrent curves) that represent actions at different levels to generate coarse proposals, while CAN refines these proposals by a multi-scale cascaded 1D-convolutional anchor network.

\textbf{Temporal Action Localization (TAL)}: For TAL task, we follow the standard ``detection by classification" framework, i.e., first generate proposals by our temporal action proposal system and then classify proposals with two-stream P3D classifier.

\textbf{Dense-Captioning Events in Videos (DCEV)}: For DCEV task, we firstly adopt our temporal action proposal system mentioned above to localize temporal proposals of interest in video, and then generate the descriptions for each proposal. Specifically, RNNs encode a given video and its detected attributes into a fixed dimensional vector, and then decode it to the target output sentence. Moreover, we extend the attributes-based CNNs plus RNNs model with policy gradient optimization and retrieval mechanism to further boost video captioning performance.

\textbf{Trimmed Action Recognition (TAR)}: We investigate and exploit multiple spatio-temporal clues for trimmed action recognition task, i.e., frame, short video clip and motion (optical flow) by leveraging 2D or 3D convolutional neural networks (CNNs). The mechanism of different quantization methods is studied as well. All activities are finally classified by late fusing the predictions from each clue.

\textbf{Spatio-temporal Action Localization (SAL)}: Our system for SAL includes two main components: i.e., Recurrent Tubelet Proposal (RTP) networks and Recurrent Tubelet Recognition (RTR) networks. The RTP initializes action proposals of the start frame through a Region Proposal Network on the feature map and then estimates the movements of proposals in the next frame in a recurrent manner. The action proposals of different frames are linked to form the tubelet proposals. The RTR capitalizes on a multi-channel architecture, where in each channel, a tubelet proposal is fed into a CNN plus LSTM network to recurrently recognize action in the tubelet.

\end{abstract}

\section{Introduction}
Recognizing activities in videos is a challenging task as video is an information-intensive media with complex variations. In particular, an activity may be represented by different clues including frame, short video clip, motion (optical flow) and long video clip. In this work, we aim at investigating these multiple clues to activity classification in trimmed videos, which consist of a diverse range of human focused actions.

Besides detecting actions in manually trimmed short video, researchers tend to develop techniques for detecting actions in untrimmed long videos in the wild. This trend motivates another challenging task---temporal action localization which aims to localize action in untrimmed long videos. We also explore this task in this work. However, most of the natural videos in the real world are untrimmed videos with complex activities and unrelated background/context information, making it hard to directly localize and recognize activities in them. One possible solution is to quickly localize temporal chunks in untrimmed videos containing human activities of interest and then conduct activity recognition over these temporal chunks, which largely simplifies the activity recognition for untrimmed videos. Generating such temporal action chunks in untrimmed videos is known as the task of temporal action proposals, which is also exploited here.

Furthermore, action detection with accurate spatio-temporal location in videos, i.e., spatio-temporal action localization, is another challenging task in video understanding and we study this task in this work. Compared to temporal action localization which temporally localizes actions, this task is more difficult due to the complex variations and large spatio-temporal search space.

In addition to the above four tasks tailored to activity which is usually the name of action/event in videos, the task of dense-captioning events in videos is explored here which goes beyond activities by describing numerous events within untrimmed videos with multiple natural sentences.

The remaining sections are organized as follows. Section 2 presents all the features which will be adopted in our systems, while Section 3 details the feature quantization strategies. Then the descriptions and empirical evaluations of our systems for five tasks are provided in Section 4-8 respectively, followed by the conclusions in Section 9.

\section{Video Representations}
We extract the video representations from multiple clues including frame, short clip, motion and long clip.

\textbf{Frame.}
To extract frame-level representations from video, we uniformly sample 25 frames for each video/proposal, and then use pre-trained 2D CNNs as frame-level feature extractors. We choose the most popular 2D CNNs in image classification---ResNet \cite{he2015deep}.

\begin{figure}[!tb]
   \centering
   \subfigure[P3D-A]{
     \label{fig:fig1:a}
     \includegraphics[width=0.15\textwidth]{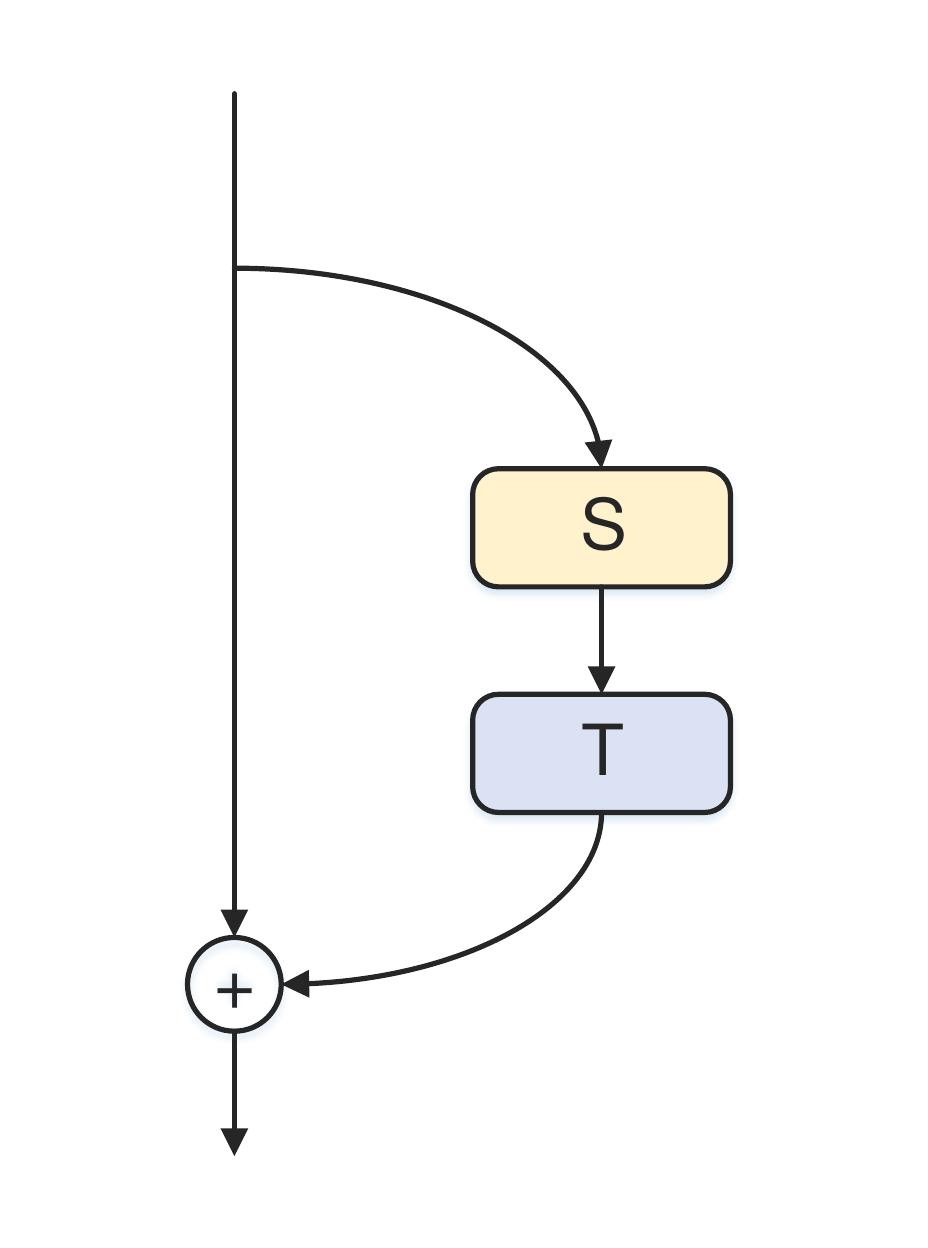}}
   \subfigure[P3D-B]{
     \label{fig:fig1:b}
     \includegraphics[width=0.15\textwidth]{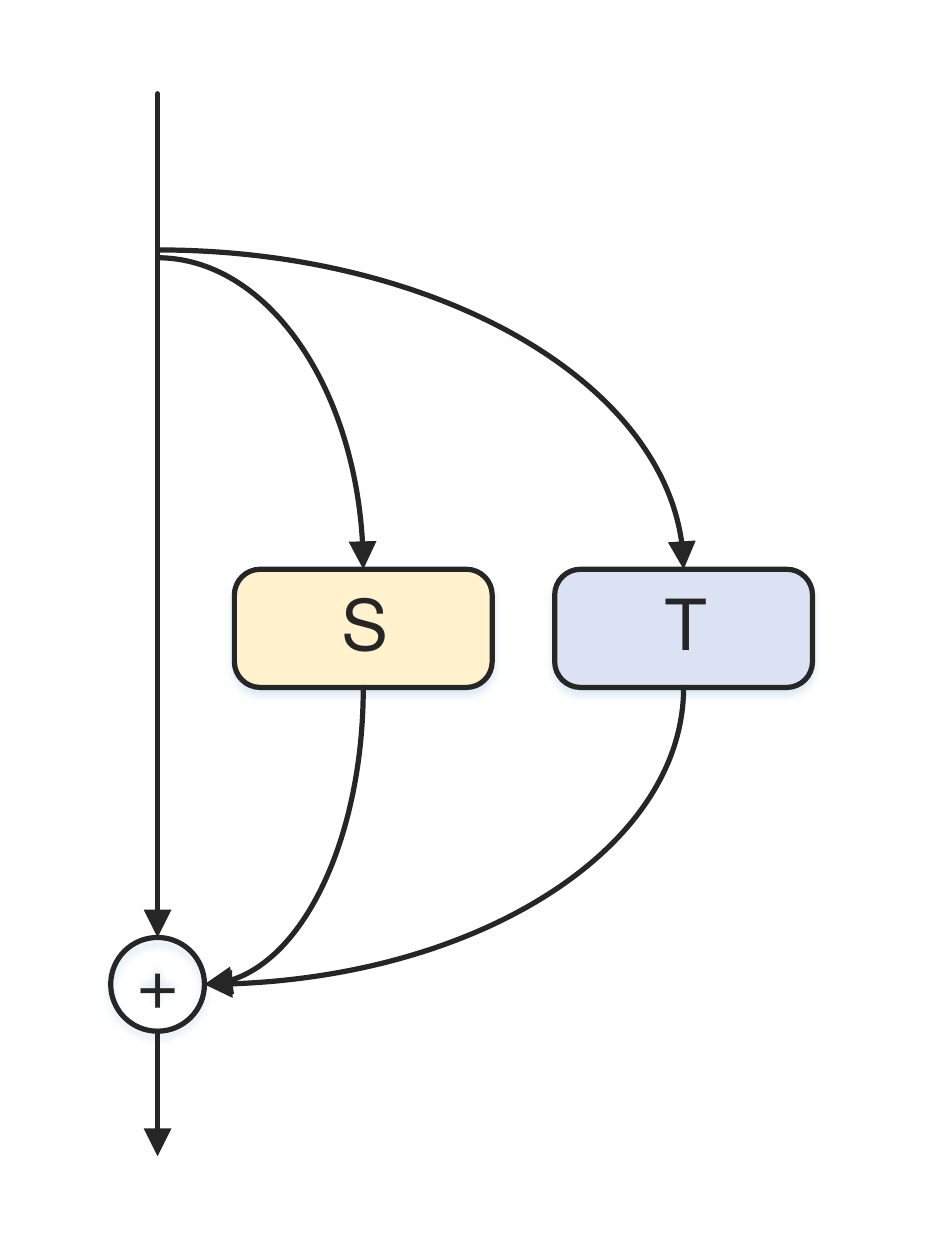}}
   \subfigure[P3D-C]{
     \label{fig:fig1:c}
     \includegraphics[width=0.15\textwidth]{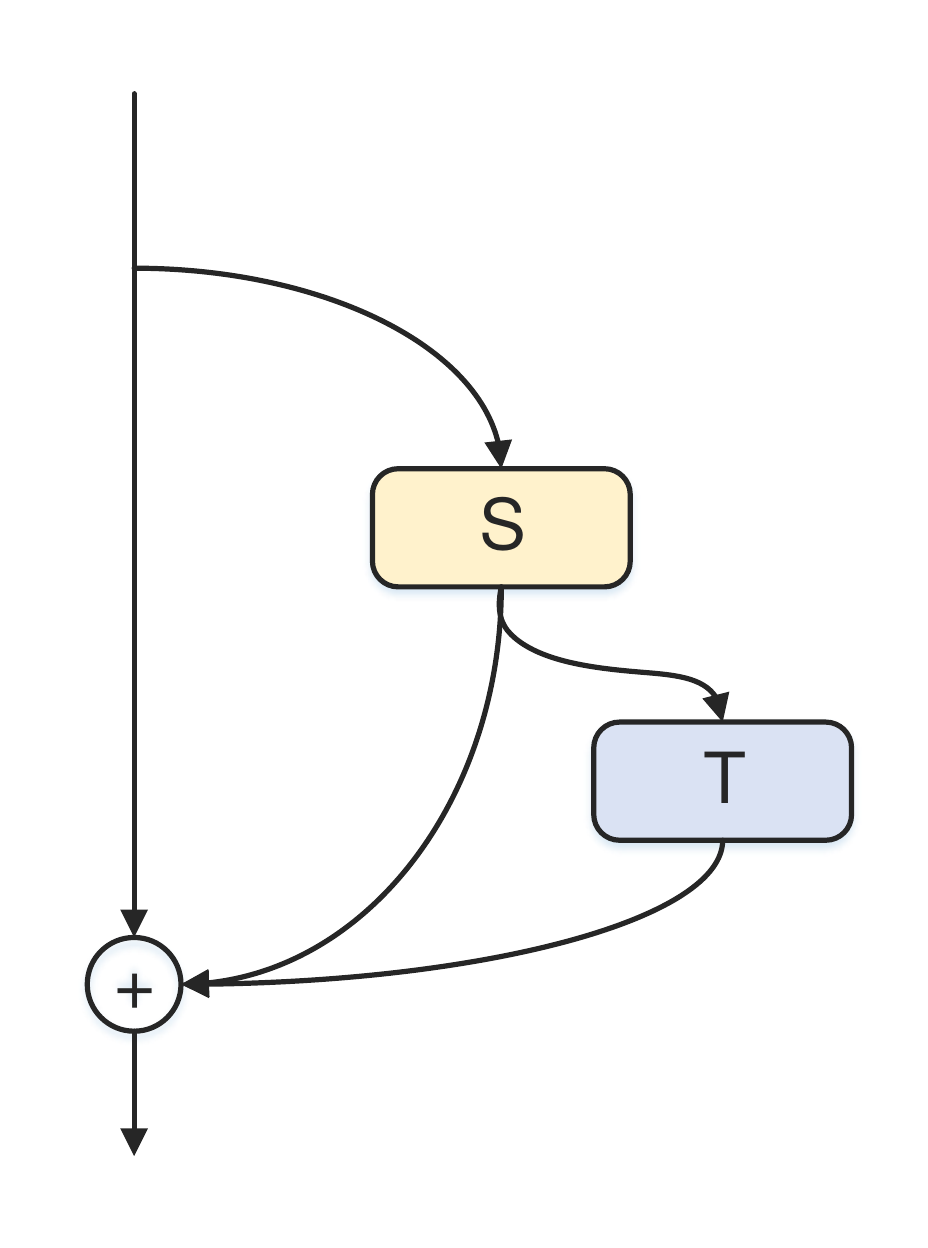}}
   \caption{\small Three Pseudo-3D blocks.}
   \label{fig:figP3D}
\end{figure}

\textbf{Short Clip.}
In addition to frame, we take the inspiration from the most popular 3D CNN architecture C3D \cite{tran2014learning} and devise a novel Pseudo-3D Residual Net (P3D ResNet) architecture \cite{qiu2017learning} to learn spatio-temporal video clip representation in deep networks. Particularly, we develop variants of bottleneck building blocks to combine 2D spatial and 1D temporal convolutions, as shown in Figure \ref{fig:figP3D}. The whole P3D ResNet is then constructed by integrating Pseudo-3D blocks into a residual learning framework at different placements. We fix the sample rate as 25 per video.

\textbf{Motion.}
To model the change of consecutive frames, we apply another CNNs to optical flow ``image,'' which can extract motion features between consecutive frames. When extracting motion features, we follow the setting of \cite{wang2015towards}, which fed optical flow images, consisting of two-direction optical flow from multiple consecutive frames, into ResNet/P3D ResNet network in each iteration. The sample rate is also set to 25 per video.

\textbf{Audio.}
Audio feature is the most global feature (though entire video) in our system. Although audio feature itself can not get very good result for action recognition, but it can be seen as powerful additional feature, since some specific actions are highly related to audio information. Here we utilize MFCC to extract audio features.

\section{Feature Quantization}
In this section, we describe two quantization methods to generate video-level/clip-level representations.

\textbf{Average Pooling.}
Average pooling is the most common method to extract video-level features from consecutive frames, short clips and long clips. For a set of frame-level or clip-level features $F=\{f_1, f_2, ..., f_N\}$, the video-level representations are produced by simply averaging all the features in the set:
\begin{equation}\label{Eq:Eq1}
\begin{array}{l}
R_{pooling}=\frac{1}{N}\sum\limits_{i:f_i \in F}{f_i}
\end{array},
\end{equation}
where $R_{pooling}$ denotes the final representations.

\textbf{Compact Bilinear Pooling.}
Moreover, we utilize Compact Bilinear Pooling (CBP) \cite{gao2016compact} to produce highly discriminative clip-level representation by capturing the pairwise correlations and modeling interactions between spatial locations within this clip. In particular, given a clip-level feature $F_t \in \mathbb{R}^{W\times H\times D}$ ($W$, $H$ and $D$ are the width, height and channel numbers), Compact Bilinear Pooling is performed by kernelized feature comparison, which is defined as
\begin{equation} \label{Eq:cbp} \small
\begin{aligned}
R_{CBP}=\sum_{j=1}^{S}\sum_{k=1}^{S}{}\left \langle \phi (F_{t,j}), \phi (F_{t,k}) \right \rangle~,
\end{aligned}
\end{equation}
where $S=W\times H$ is the size of the feature map, $F_{t,j}$ is the region-level feature of $j$-th spatial location in $F_t$, $\phi (\cdot)$ is a low dimensional projection function, and $\left \langle \cdot \right \rangle$ is the second order polynomial kernel.

\begin{figure*}[!tb]
\centering {\includegraphics[width=0.9\textwidth]{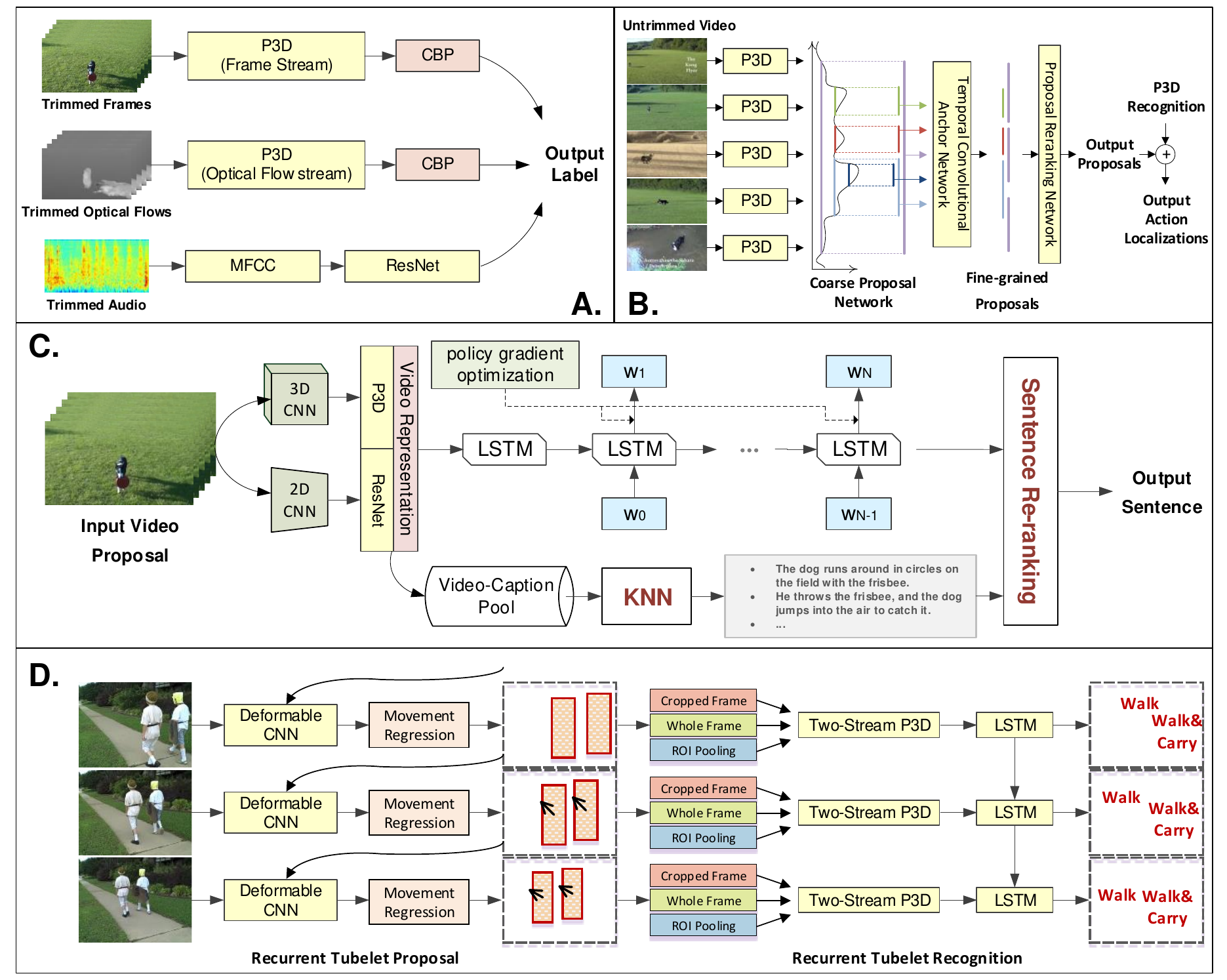}}
\caption{Frameworks of our proposed (a) trimmed action recognition system, (b) temporal action proposals system, (c) dense-captioning events in videos system, and (d) spatio-temporal action localization system.}
\label{fig:figall}
\end{figure*}

\section{Trimmed Action Recognition}
\subsection{System}
Our trimmed action recognition framework is shown in Figure \ref{fig:figall} (a). In general, the trimmed action recognition process is composed of three stages, i.e., multi-stream feature extraction, feature quantization and prediction generation. For deep feature extraction, we follow the multi-stream approaches in \cite{li2016action,qiumsr16,qiumsr,qiu2017deep}, which represented input video by a hierarchical structure including individual frame, short clip and consecutive frame. In addition to visual features, the most commonly used audio feature MFCC is exploited to further enrich the video representations. After extraction of raw features, different quantization and pooling methods are utilized on different features to produce global representations of each trimmed video. Finally, the predictions from different streams are linearly fused by the weights tuned on validatoin set.

\subsection{Experiment Results}
Table \ref{table:tab1} shows the performances of all the components in our trimmed action recognition system. Overall, the CBP on P3D ResNet (128-frame) achieves the highest top1 accuracy (78.47\%) and top5 accuracy (93.99\%) of single component. And by additionally apply this model on both frame and optical flow, the two-stream P3D achieves an abvious improvement, which gets top1 accuracy of 80.91\% and top5 accuracy of 94.96\%. For the final submission, we linearly fuse all the components.

\begin{table*}
\centering
\caption{Comparison of different components in our trimmed action recognition framework on Kinetics validation set for trimmed action recognition task.}
\label{table:tab1}
\begin{tabular}{l|c|c|c|c|c} \hline
~~~\textbf{Stream}&~~~\textbf{Feature}~~~&~~~\textbf{Layer}~~~&~~~\textbf{Quantization}~~~&~~~\textbf{Top1}~~~&~~~\textbf{Top5}~~~\\ \hline
~~~\multirow{2}{*}{Frame}
& ResNet &pool5 & Ave & 74.11\% & 91.51\% \\
& ResNet &res5c & CBP & 74.97\% & 91.48\%\\
\hline
~~~\multirow{3}{*}{Short Clip}
& P3D ResNet (16-frame) &pool5 & Ave & 76.22\% & 92.92\% \\
& P3D ResNet (128-frame) &pool5 & Ave & 77.94\% & 93.75\% \\
& P3D ResNet (128-frame) &res5c & CBP & 78.47\% & 93.99\%\\
\hline
~~~\multirow{3}{*}{Motion}
& P3D ResNet (16-flow) &pool5 & Ave & 64.37\% & 85.76\% \\
& P3D ResNet (128-flow) &pool5 & Ave & 69.87\% & 89.44\% \\
& P3D ResNet (128-flow) &res5c & CBP & 71.07\% & 90.00\%\\
\hline
~~~\multirow{1}{*}{Audio}
& ResNet &pool5 & Ave & 21.91\% & 38.49\% \\
\hline
~~~\multirow{1}{*}{Two-stream P3D}
& P3D ResNet (128-frame\&flow) &res5c & CBP & 80.91\% & 94.96\% \\
\hline
~~~\multirow{1}{*}{Fusion all}
&  &&  & 83.75\% & 95.95\% \\

\hline
\end{tabular}
\end{table*}

\section{Temporal Action Proposals}
\subsection{System}
Figure \ref{fig:figall} (b) shows the framework of temporal action proposals, which is mainly composed of three stages:

\textbf{Coarse Proposal Network (CPN).} In this stage, proposal candidates are generated by watershed temporal actionness grouping algorithm (TAG) based on actionness curve. Considering the diversity of action proposals, three actionness measures (namely point-wise, pair-wise and recurrent) that are complementary to each other are leveraged to produce the final actionness curve.

\textbf{Temporal Convolutional Anchor Network (CAN).} Next, we feed long proposals into our temporal convolutional anchor network for finer proposal generation. The temporal convolutional anchor network consists of multiple 1D convolution layers to generate temporal instances for proposal/background binary classification and bounding box regression.

\textbf{Proposal Reranking Network (PRN).} Given the short proposals from the coarse stage
and fine-grained proposals from the temporal convolutional anchor network, a reranking network is utilized for proposal refinement. To take video temporal structures into account, we extend the current part of proposal with its' start and end part.
The duration of start and end parts are half of the current part. The proposal is then
represented by concatenating features of each part to leverage the context information. In our experiments, the top 100 proposals are finally outputted.

\subsection{Experiment Results}
Table \ref{table:tab2} shows the action proposal AUC performances of frame/optical flow input to P3D~\cite{qiu2017learning} with different stages in our system. The two stream P3D architecture is pre-trained on Kinetics \cite{kay2017kinetics} dataset. For all the single stream runs with different stages, the setting based on all three stages combination achieves the highest AUC. For the final submission, we combine all the proposals from the two streams and then select the top 100 proposals based on their weighted ranking probabilities. The linear fusion weights are tuned on validation set.

\begin{table}
\centering
\caption{Area Under the average recall vs. average number of proposals per video Curve (AUC) of frame/flow input for P3D~\cite{qiu2017learning} network on ActivityNet validation set for temporal action proposals task.}
\label{table:tab2}
\begin{tabular}{l|c c c|c}\hline
~\textbf{Stream}&~\textbf{CPN}~&~\textbf{CAN}~&~\textbf{PRN}~&~\textbf{AUC}~\\ \hline
\multicolumn{1}{c|}{\multirow{3}{*}{\text{Frame}}}	    & $\surd$ & & & 60.27\% \\
\multicolumn{1}{c|}{} & $\surd$ & $\surd$ & & 63.20\% \\
\multicolumn{1}{c|}{} & $\surd$ & $\surd$ & $\surd$ & 64.21\% \\\hline
\multicolumn{1}{c|}{\multirow{3}{*}{\text{Optical Flow}}}	    & $\surd$ & & & 59.83\% \\
\multicolumn{1}{c|}{} & $\surd$ & $\surd$ && 63.43\% \\
\multicolumn{1}{c|}{} & $\surd$ & $\surd$ &$\surd$& 64.02\% \\\hline
~Fusion all & & & &\textbf{67.36\%} \\\hline
\end{tabular}
\end{table}

\section{Temporal Action Localization}

\subsection{System}
Without loss of generality, we follow the standard ``detection by classification" framework, i.e., first generate proposals by temporal action proposals system and then classify proposals. The action classifier is trained with the above trimmed action recognition system (i.e., two-stream P3D) over the 200 categories on ActivityNet dataset \cite{caba2015activitynet}.

\begin{table}
\centering
\caption{Performance comparison of different methods on ActivityNet validation set for temporal action localization task. Results are evaluated by mAP with different IoU thresholds and average mAP of IoU threshold from 0.5 to 0.95 with step 0.05.}
\label{table:tab3}
\begin{tabular}{l|c c c|c}\hline
~\textbf{mAP}&~\textbf{0.5}~&~\textbf{0.75}~&~\textbf{0.95}~&~\textbf{Avg mAP}~\\ \hline
\text{Shou et al.~\cite{Shou:CVPR17}}    & 43.83 & 25.88 & 0.21 & 22.77\\ \hline
\text{Xiong et al.~\cite{Xiong:ARXIV17}}	  & 39.12 & 23.48 & 5.49 & 23.98 \\ \hline
\text{Lin et al.~\cite{Lin:ARXIV17}}     & 48.99 & 32.91 & 7.87 & 32.26 \\ \hline
\text{Ours}  & 51.40 & 33.61 & 8.13 & 34.22  \\ \hline
\end{tabular}
\end{table}

\subsection{Experiment Results}
Table~\ref{table:tab3} shows the action localization mAP performance of our approach and baselines on validation set. Our approach consistently outperforms other state-of-the-art approaches in different IoU threshold and achieves 34.22\% average mAP on validation set.

\begin{table*}
\centering
\caption{Performance of our dense-captioning events in videos system on ActivityNet captions validation set, where B@$N$, M, R and C are short for BLEU@$N$, METEOR, ROUGE-L and CIDEr-D scores. All values are reported as percentage (\%).}
\label{table:tabdense}
\begin{tabular}{l|c|c|c|c|c|c|c}\hline
~~~\textbf{Model}&~~~~~\textbf{B@1}~~~~&~~~~\textbf{B@2}~~~~&~~~~\textbf{B@3}~~~~&~~~~\textbf{B@4}~~~~
&~~~~\textbf{M}~~~~&~~~~\textbf{R}~~~~&~~~~\textbf{C}~~~~\\ \hline
~~~\textbf{{LSTM-A}$_{3}$}                               & 13.78    & 7.12    & 3.53    & 1.72    & 7.61    & 13.30    & 27.07 \\
~~~\textbf{{LSTM-A}$_{3}$ + policy gradient}             & 11.65    & 6.05    & 3.02    & 1.34    & 8.28    & 12.63    & 14.62 \\
~~~\textbf{{LSTM-A}$_{3}$ + policy gradient + retrieval} & 11.91    & 6.13    & 3.04    & 1.35    & 8.30    & 12.65    & 15.61 \\\hline
\end{tabular}
\end{table*}

\section{Dense-Captioning Events in Videos}
\subsection{System}
The main goal of dense-captioning events in videos is jointly localizing temporal proposals of interest in videos and then generating the descriptions for each proposal/video clip. Hence we firstly leverage the temporal action proposal system described above in Section 5 to localize temporal proposals of events in videos (2 proposals for each video). Then, given each temporal proposal (i.e., video segment describing one event), our dense-captioning system runs two different video captioning modules in parallel---the generative module for generating caption via the LSTM-based sequence learning model, and the retrieval module which can directly copy sentences from other visually similar video segments through KNN. Specifically, the generative module with LSTM is inspired from the recent successes of probabilistic sequence models leveraged in vision and language tasks (e.g., image captioning \cite{vinyals2015show,yao2017novel}, video captioning \cite{pan2016jointly,pan2017seeing,pan2017video}, video generation from captions \cite{pan2017to} and dense video captioning \cite{li2018jointly,yao2017msr}). We mainly utilize the third design LSTM-A$_3$ in \cite{yao2017boosting} which firstly encodes attribute representations into LSTM and then transforms video representations into LSTM at the second time step is adopted as the basic architecture. Note that we employ the policy gradient optimization method with reinforcement learning \cite{Rennie:2016SCST} to boost the video captioning performances specific to METEOR metric. For the retrieval module, we utilize KNN to find the visually similar video segments based on the extracted video representations. The captions associated with the top similar video segments are regarded as sentence candidates in retrieval module. In the experiment, we mainly choose the top 300 nearest neighbors for generating sentence candidates. Finally, a sentence re-ranking module is exploited to rank and select the final most consensus caption from the two parallel video captioning modules by considering the lexical similarity among all the sentence candidates. The overall architecture of our dense-captioning system is shown in Figure \ref{fig:figall} (c).

\subsection{Experiment Results}
Table \ref{table:tabdense} shows the performances of our proposed dense-captioning events in videos system. Here we compare three variants
derived from our proposed model. In particular, by additionally incorporating the policy gradient optimization scheme into the basic LSTM-A$_{3}$ architecture, we can clearly observe the performance boost in METEOR. Moreover, our dense-captioning model ({LSTM-A}$_{3}$ + policy gradient + retrieval) is further improved by injecting the sentence candidates from retrieval module in METEOR.

\section{Spatio-temporal Action Localization}

\subsection{System}
Figure \ref{fig:figall} (d) shows the framework of spatio-temporal action localization, which includes two main components:

\textbf{Recurrent Tubelet Proposal (RTP) networks.}
The Recurrent Tubelet Proposal networks produces action proposals in a recurrent manner. Specifically, it initializes action proposals of the start frame through a Region Proposal Network (RPN) \cite{ren2015faster} on the feature map. Then the movement of each proposal in the next frame is estimated from three inputs: feature maps of both current and next frames, and the proposal in current frame. Simultaneously, a sibling proposal classifier is utilized to infer the actionness of the proposal. To form the tubelet proposals, action proposals in two consecutive frames are linked by taking both their actionness and overlap ratio into account, followed by the temporal trimming on tubelet.

\textbf{Recurrent Tubelet Recognition (RTR) networks.}
The Recurrent Tubelet Recognition networks capitalizes on a multi-channel architecture for tubelet proposal recognition. For each proposal, we extract three different semantic-level features, i.e., the features on proposal-cropped image, the features with RoI pooling on the proposal, and the features on whole frame. These features implicitly encode the spatial context and scene information, which could enhance the recognition capability on specific categories. After that, each of them is fed into a LSTM to model the temporal dynamics for tubelet recognition.

\subsection{Experiment Results}
We construct our RTP based on \cite{dai2017deformable}, which is mainly trained with the single RGB frames. For RTR, we extract the region representations with RoI pooling from multiple clues including frame, clip and motion. Table \ref{table:AVA} shows the performances of all the components in our RTR. Overall, the P3D ResNet trained on clips (128 frames) achieves the highest frame-mAP (19.40\%) of single component. For the final submission, all the components are linearly fused using the weights tuned on validation set. The final mAP on validation set is 22.20\%.

\begin{table}[]
\centering
\caption{Comparison of different components in our RTR on AVA validation set for spatio-temporal action localization task.}
\label{table:AVA}
\begin{tabular}{l|c|c}
\hline
\textbf{Stream}         & \textbf{Feature} & \textbf{mAP@IoU=0.5} \\ \hline
Frame          & ResNet           & 13.68                \\ \hline
Short Clip     & P3D ResNet (16-frame)      & 19.12                \\ \hline
Short Clip     & P3D ResNet (128-frame)       & 19.40                \\ \hline
Flow                 & P3D ResNet (16-frame)      & 15.20                \\ \hline
Fusion                  & -                & \textbf{22.20}       \\ \hline
\end{tabular}
\end{table}

\section{Conclusion}
In ActivityNet Challenge 2018, we mainly focused on multiple visual features, different strategies of feature quantization and video captioning from different dimensions. Our future works include more in-depth studies of how fusion weights of different clues could be determined to boost the action recognition/temporal action proposals/temporal action localization/spatio-temporal action localization performance and how to generate open-vocabulary sentences for events in videos.

{
\bibliographystyle{ieee}
\bibliography{egbib}
}

\end{document}